\RequirePackage{fix-cm}
\documentclass[smallextended]{svjour3}       
\smartqed  
\usepackage{graphicx}
\usepackage{multirow}
\usepackage{amsmath}
\usepackage{threeparttable}
\usepackage{array}
\usepackage{amssymb}
\usepackage{threeparttable}
\usepackage{makecell}
\usepackage{url}
\usepackage{subfigure}
\usepackage{booktabs}
\usepackage{graphicx}
\usepackage{colortbl}
\usepackage{mathrsfs}
\usepackage{CJKutf8}
\usepackage{xfrac}
\usepackage{enumitem}
\usepackage{comment}

\begin{document}

\title{Annotation of Chinese Predicate Heads and Relevant Elements
}


\author{Yanping Chen  \and
        Wenfan Jin \and
        Yongbin Qin \and
        Ruizhang Huang \and
        Qinhua Zheng \and
        Ping Chen 
}


\institute{F. Yanping Chen \at
              Huaxi District, Guiyang City, Guizhou Province, 550025, P.R. China \\
              Tel.: +86-18798725895\\
              \email{ypench@gmail.com}           
           \and
           S. Wenfan Jin \at
              Guizhou University, Guiyang, 550025
			  \and
			 T. Yongbin Qin \at
			 Guizhou University, Guiyang, 550025 
			  \and
           F. Ruizhang Huang \at
           Guizhou University, Guiyang, 550025
           \and
           F. Qinghua Zheng \at
           Xi'an Jiaotong University, Xi'an, China
           \and
           S. Ping Chen \at
           University of Massachusetts Boston, Boston, USA
}

\date{Received: date / Accepted: date}

\maketitle

\begin{abstract}
A predicate head is a verbal expression that plays a role as the structural center of a sentence. Identifying predicate heads is critical to understanding a sentence. It plays the leading role in organizing the relevant syntactic elements in a sentence, including subject elements, adverbial elements, etc. For some languages, such as English, word morphologies are valuable for identifying predicate heads. However, Chinese offers no morphological information to indicate words’ grammatical roles. A Chinese sentence often contains several verbal expressions; identifying the expression that plays the role of the predicate head is not an easy task. Furthermore, Chinese sentences are inattentive to structure and provide no delimitation between words. Therefore, identifying Chinese predicate heads involves significant challenges.

In Chinese information extraction, little work has been performed in predicate head recognition. No generally accepted evaluation dataset supports work in this important area. This paper presents the first attempt to develop an annotation guideline for Chinese predicate heads and their relevant syntactic elements. This annotation guideline emphasizes the role of the predicate as the structural center of a sentence. The design of relevant syntactic element annotation also follows this principle. Many considerations are proposed to achieve this goal, e.g., patterns of predicate heads, a flattened annotation structure, and a simpler syntactic unit type. Based on the proposed annotation guideline, more than 1,500 documents were manually annotated. The corpus will be available online for public access. With this guideline and annotated corpus, our goal is to broadly impact and advance the research in the area of Chinese information extraction and to provide the research community with a critical resource that has been lacking for a long time \footnote{The corpus is available at:\url{https://www.editorialmanager.com/lrev/default.aspx}}. 
   
\end{abstract}

\section{Introduction}
\label{intro}
Predicates are grammatical components of sentences. Chinese information extraction has two coexistent definitions for predicates. In the first definition, a predicate is a verbal expression that represents an action or a change in state. This definition allows multiple predicates in a sentence. The second definition (referred to as the predicate head), in addition to expressing a verbal meaning, requires the verbal expression to be the structural center of a sentence.

A sentence often encapsulates several relevant syntactic elements. For ease of understanding, having a semantic focus is helpful for recipients to capture the sentence meaning. The predicate head acts as the structural center of a sentence, which organizes other syntactic elements. Therefore, defining predicate heads as the structural centers of sentences has great practical significance and can help extract syntactic or semantic information from sentences. In this paper, we adopt the definition: {\bf a predicate head is a verbal expression that acts as the structural and semantic center of a sentence}. This definition also has a solid theoretical foundation. This definition aligns well with theories in linguistics psychology \cite{Ref1,Ref2}.

A predicate head is usually a verbal expression representing an occurrence of an “animated concept”, e.g., a behavior in living things, a movement in positions or a change in states. In natural language processing, a similar concept is the “named entity”, coined in the Sixth Message Understanding Conference (MUC)\cite{Ref3}. An entity is an object or set of objects in the world \cite{Ref4}. An occurrence of an entity in a sentence is called an “entity mention”.

The task of named entity recognition has been extensively studied. Conversely, the task of predicate head recognition has received very little attention. We think that there are three reasons for this phenomenon. First, most entity names are open category words. Automatically recognizing these entities is a challenging task. On the other hand, the number of verbs is relatively stable, especially in English. For example, the Oxford English Dictionary accepted 235 new word entries in 2018, where only 24 entries are verbs (e.g., mansplain, self-administer, etc.). Second, it is considered that recognizing verbal expressions is easier, because they have a simple structure (in English), and many morphologies are helpful for identifying them. Third, predicate heads are usually considered verbs. Part-of-speech tagging can address this problem.

However, these factors do not apply when annotating Chinese predicate heads. First, reduplicated method and idioms are widely used to generate new verbal expressions, which are used directly as predicate heads of sentences. Second, in Chinese, no morphological cue is available to indicate the syntactic or semantic roles of verbal expressions. Third, the part-of-speech tagging is not effective to recognize predicate heads, because they are usually segmented into smaller words in Chinese word segmentation. Therefore, annotating Chinese predicate heads is both necessary and challenging.

\section{Challenges for Annotating Predicate Head}
\label{sec:challenges}
To identify a predicate head involves two steps: identifying verbal expressions in a sentence, and collecting the verbal expression that acts as the semantic focus of the sentence. Both are challenging issues in Chinese natural language processing.

Before discussing the details of the challenge, here is one example about Chinese predicate heads. This example raises several issues about the annotation of Chinese predicate heads.

\begin{enumerate}
\item \begin{CJK}{UTF8}{gbsn}被告人陈某某因家庭矛盾{\bf 迁怒}岳父滕某某。2015年6月29日凌晨，陈某某{\bf 谎称}购买房屋，将其{\bf 骗}至其新房南侧桥上，两人{\bf 发生}争执并互相{\bf 厮打}。陈某某持刀{\bf 捅刺}滕某某，用砖头多次{\bf 击打}其头部，并将其头部{\bf 撞向}地面，致其死亡。陈某某驾驶电动三轮车{\bf 抛}尸至大桥下的河中\end{CJK}\footnote{Due to family conflicts, defendant Chen Moumou angered his father-in-law Teng Moumou. In the early morning of June 29/2015, Chen lied about buying a house. Chen lured Teng to the south side of his new house. They argued and fought with each other. Chen Moumou held a knife and stabbed Teng Moumou, hit his head multiple times with bricks, hit his head against the ground, and caused him to die. Chen Moumou drove an electric tricycle and threw the body in a river.}
\end{enumerate}

In this example, several properties can be elucidated:
\begin{itemize}
\item[1)] This example shows that annotated verbal expressions are important for understanding the story. They plot the outline of a story.

\item[2)] Each predicate head plays a central role for organizing the linguistic units in a sentence. For example, in the sentence \begin{CJK}{UTF8}{gbsn}“2015年6月29日凌晨, 陈某某谎称购买房屋”, \end{CJK}the phrase \begin{CJK}{UTF8}{gbsn}“2015年6月29日凌晨”\end{CJK} suggests the time when an action occurs. \begin{CJK}{UTF8}{gbsn}“陈某某”\end{CJK} is the subject who implements the action ({\bf \begin{CJK}{UTF8}{gbsn}谎称\end{CJK}}, which means lie). The content of {\bf \begin{CJK}{UTF8}{gbsn}“谎称”\end{CJK}} is \begin{CJK}{UTF8}{gbsn}“购买房屋”\end{CJK}.

\item[3)]Not all verbal expressions are regarded as predicate heads. For example, in the sentence \begin{CJK}{UTF8}{gbsn}“陈某某谎称购买房屋”\end{CJK}, \begin{CJK}{UTF8}{gbsn}“谎称”\end{CJK} and \begin{CJK}{UTF8}{gbsn}“购买”\end{CJK} are verbal expressions. In this sentence, \begin{CJK}{UTF8}{gbsn}“谎称”\end{CJK} is the predicate, while \begin{CJK}{UTF8}{gbsn}“购买房屋”\end{CJK} is a noun phrase.

\item[4)]The sentence \begin{CJK}{UTF8}{gbsn}“陈某某驾驶电动三轮车抛尸至大桥下的河中”\end{CJK} contains two clauses: \begin{CJK}{UTF8}{gbsn}“驾驶电动三轮车” \end{CJK}(drive an electro-tricycle) and \begin{CJK}{UTF8}{gbsn}“抛尸抛至大桥下的河中”\end{CJK} (throw the body into the river under the bridge). \begin{CJK}{UTF8}{gbsn}“驾驶” \end{CJK}and \begin{CJK}{UTF8}{gbsn}“抛”\end{CJK} are two verbs. However, without word morphological information, it is difficult to determine the predicate head.

\item[5)]Another important characteristic is shown in the second and third sentences, where each sentence is composed of several clauses divided by commas. Some clauses share the same subject. For example, the third sentence contains three verbs: \begin{CJK}{UTF8}{gbsn}“捅刺”, “击打”, “撞向”\end{CJK}. These verbs are equally important.

\item[6)] The clause \begin{CJK}{UTF8}{gbsn}“两人发生争执并互相厮打”\end{CJK} has two verbs, \begin{CJK}{UTF8}{gbsn}“发生”\end{CJK} and \begin{CJK}{UTF8}{gbsn}“厮打”\end{CJK}. Because of the conjunction \begin{CJK}{UTF8}{gbsn}“并”\end{CJK}, determining the predicate head is ambiguous.
\end{itemize}

The above examples show that annotating Chinese predicate heads is not an easy task. It deserves careful consideration and faces significant research challenges. In summary, the followings give six challenges when annotating Chinese predicate heads.

{\bf Segmentation Ambiguity:} A Chinese sentence is written character by character without delimitation between words. Because Chinese has tens of thousands of characters, and almost every character can be simultaneously seen as a word or as a morpheme in a sentence, segmenting verbal expressions from a sentence suffers from serious segmentation ambiguities. These ambiguities can be classified into two categories: overlapping ambiguity and combinational ambiguity (Chen et al., 2016). In overlapping ambiguity, a character string contains verbs that are overlapped. For example, in the phrase \begin{CJK}{UTF8}{gbsn}“结合成分子”\end{CJK}, the overlapping words are \begin{CJK}{UTF8}{gbsn}“结合”\end{CJK} (combine) and \begin{CJK}{UTF8}{gbsn}“合成”\end{CJK} (synthetize). Combinational ambiguity is caused by the fact that every Chinese character can be either a morpheme or a word. For example,\begin{CJK}{UTF8}{gbsn} “合成”\end{CJK} is a word. This word can also be divided as \begin{CJK}{UTF8}{gbsn}“合/ 成”\end{CJK} (bear/ join). Without complete semantic information about the sentence, distinguishing the morpheme and the word is often difficult.

{\bf Unknown Words:} Generating new words with existing words is an important word formation rule in Chinese. Many verbal expressions are dynamically generated by rules. Listing all of these combined expressions in advance is impossible. In field of natural language processing, these expressions are referred as ``unknown words''. In current approaches, combined verbal expressions are often segmented into pieces, a process that completely ignores the syntactic roles of verbal expressions. For example, the word \begin{CJK}{UTF8}{gbsn}“打砸抢”\end{CJK} is composed of three verbs \begin{CJK}{UTF8}{gbsn}“打/砸/抢”\end{CJK} (beat/ smash/ loot), which generates the new meaning, “behaviors to create chaos”.  The problem is that many compound words are widely used but are not registered in any dictionary, e.g., \begin{CJK}{UTF8}{gbsn}“抬头望去”\end{CJK} (look up) and \begin{CJK}{UTF8}{gbsn}“开发建设”\end{CJK} (development and construction). Both are often segmented as \begin{CJK}{UTF8}{gbsn}“抬头/望去”\end{CJK} and \begin{CJK}{UTF8}{gbsn}“开发/建设”\end{CJK}. Because these combined verbal expressions often act in independent syntactic roles in a sentence, segmenting them into smaller units is not feasible for analysing sentence structure.

{\bf Reduplicated Structure:} Chinese often duplicates characters and words to generate compound words, e.g., \begin{CJK}{UTF8}{gbsn} AA, AAB, ABB, AABB, A 里 AB, A 不 AB, and ABAB (e.g., “走走”, “跑一跑”, “洗洗澡”, “勾勾搭搭”, “慌里慌张”, and “比划比划”\end{CJK}). Commonly, verbs with reduplicated structures are used to emphasize a semantic aspect. For example, \begin{CJK}{UTF8}{gbsn}“跑一跑” and “洗洗澡” \end{CJK}imply a relaxed behavior. This formation rule can generate nonenumerable compound words that are impossible to register in a lexicon. Many of these compound words are also seen as unknown words. Even they act as an independent syntactic component. Current toolkits still segment them into pieces, completely ignoring their syntactic role.

{\bf Little Morphology:} In inflected languages, such as English, morphemes are helpful to distinguish a word’s grammatical, syntactic, or semantic role. Chinese verbs are usually multi-categorical in terms of part of speech, but no morphology indicates their verbal usages. For example, the word \begin{CJK}{UTF8}{gbsn}“打”\end{CJK} can represent a quantifier (a dozen), a verb (strike), a preposition (from) or a noun (fight), etc. When a verbal meaning is expressed, it refers to “strike”, “strikes”, “struck”, “stricken” and “striking”. Furthermore, a Chinese sentence often contains several verbs, each of these verbs can be handled as a predicate head or as an adverbial phrase. The lack of morphology makes distinguishing the syntactic role between them difficult. Therefore, when a sentence contains several verbal expressions, the predicate heads are difficult to identify.

{\bf Ambiguity of Sentence Boundary:} Finding the boundaries of Chinese sentences is also a challenging task because a comma is often ambiguously used to segment sentences or clauses \cite{Ref5}. Many educated native Chinese speakers often use commas as sentence boundaries. Researchers have shown that the performance of information extraction is heavily influenced by sentence boundary criterion \cite{Ref6}. Because a predicate head is defined as the semantic focus of a sentence, annotating predicate heads is heavily influenced by the boundary ambiguity problem.

\begin{figure*}  
\includegraphics[width=.8\linewidth]{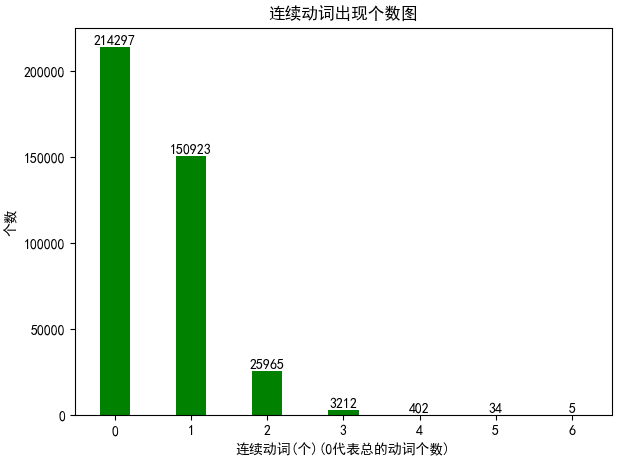}  
\caption{The distribution of adjacent verb number}
\label{fig:framework}  
\end{figure*}

{\bf Inattention to Structure: }The Chinese language is an ancient hieroglyphic in which sentences are inattentive to structure. In Chinese, a sentence often contains many successive verbs to express related actions. Examples include \begin{CJK}{UTF8}{gbsn}“二被告人/ 商量/ 决定/ 寻找/ 机会/ 杀死/ 张某”\end{CJK} (“The two defendants/ discuss/ decide/ find/ change/ kill/ Zhang Mou”), \begin{CJK}{UTF8}{gbsn}“抬头/ 望去”\end{CJK} (look over/see), and \begin{CJK}{UTF8}{gbsn}“驱车/ 行驶”\end{CJK} (drive/travel). In one month of the People’s Daily corpus \cite{Ref7}, the number of adjacent verbs extends from 2 to 6. Statistical information about the length of successive verbs is shown in Figure \ref{fig:framework}. Here is an example of a multiverbal sentence with 6 successive verbs manually labeled:\begin{CJK}{UTF8}{gbsn} “在/p 大/a 变革/vn 中/f 塑造/v 开掘/v 出/v 能/v 映照/v 出/v 时代/n 和/c 历史/n 的/u 人物/n ,/w 事件/n, /w”\end{CJK}.

In addition to the above six characteristics, there are two factors influencing the task of Chinese predicate head annotation. First, in the field of Chinese information extraction, little work has been done for annotating Chinese predicates. At current, there is no annotation guideline or corpus in this field and research community. Second, the characteristics of Chinese predicate heads indicate that recognizing them requires the modeling of high-order dependencies, in which the global features of a sentence are more important. Because current algorithms (e.g., the hidden Markov model (HMM) or the conditional random field (CRF)) often assume one-order Markov dependency on an input sequence, they are too weak to capture high-order dependency information \cite{Ref8}. The Long Short-Term Memory (LSTM) depends on a cell for remembering long dependent information in a sentence \cite{Ref9}. However, the information is deteriorated when the dependent distance is longer.

\subsection{Application of Predicate Heads}
\label{sec:application}

A predicate head plays an important syntactic role in a sentence and is often parsed as the root of a parsing tree. It organizes linguistic units in a sentence, which is the key to understanding sentences. Because many semantic relationships between sentences (e.g., causal relations) are expressed through predicate heads, identifying predicate heads is beneficial for many natural language processing (NLP) tasks. Here are a few important applications of predicate heads.

\begin{enumerate}

\item {\bf Parsing}: Because Chinese language is an ancient hieroglyphic, it is inattentive to structure. Parsing Chinese sentences is difficult. Current methods of parsing Chinese sentences heavily depend on the outputs of Chinese word segmentation and POS-tagging. These methods are far from perfect. Because word segmentation and POS-tagging often serve as foundational NLP tasks, they may cause a cascading failure, which influences the parsing tasks. On the other hand, the task of identifying predicate heads belongs to the area of information extraction, which extracts designated linguistic units directly and avoids the influence caused by other tasks. As discussed above, a predicate head is the center of a sentence. It is often the root of a parsing tree. Therefore, predicate heads are valuable for the sentence parsing task.

\item {\bf Automatic text summarization}: The task of summarization relies on accurately extracting the main concepts of a document or a set of documents \cite{Ref10}. Many features have been explored to support text summarization, e.g., word frequency or key phrases and word positions in a text. Because predicate heads are the semantic focus of sentences, they are effective for capturing the core ideal of sentences. Supported by predicate head recognition, text summarization can be implemented at different granularities. At the sentence level, extra information can be erased without changing the main sentence semantics. At the document level, an outline of a document can be drawn by selecting predicate heads that are closely related to the topic of a document. Recognizing predicate heads is beneficial for improving the fluency and coherence of summarization.

\item {\bf Knowledge graph construction}:  Knowledge graphs, such as those in Yago \cite{Ref11} and Freebase \cite{Ref12}, are widely used to represent common knowledge about a specific field. They provide a unified framework to organize semantic information. These graphs merge diverse and heterogeneous data with high scalability. They often use an extendible definition to support the scalability and extensibility of data structure. In these graphs, nodes are commonly defined as named entities. Edges denote semantic relationships between nodes. These graphs can be seen as “static” graphs, because verbal expressions are rarely adopted as nodes. On the other hand, predicate heads and semantic relationships between them (e.g., sequential or causal relationships) usually represent behavioral information. Building a “dynamic” knowledge graph with these semantic elements is desirable, which will be useful for representing narrative knowledge such as stories or events.

\item {\bf Chinese word segmentation}: Segmentation is the most fundamental task for Chinese NLP. There are two main obstacles to implementing this task: segmentation ambiguity and out-of-vocabulary (OOV) problems \cite{Ref13}. Currently, Chinese word segmentation is modeled as a sequence labeling task, in which sequence models (e.g., HMM, CRF, and LSTM) are used to find the most possible label sequence. This approach is weak for capturing the global features of a sentence. Because each predicate head plays a central role in organizing linguistic units in a sentence, identifying predicate heads in advance is helpful for supporting Chinese word segmentation.

\end{enumerate}

In this paper, we present an annotation guideline for Chinese predicate heads. The rest of this paper is organized as follows. Section \ref{sec:related-work} discusses the related work about the annotation of predicate heads. Section \ref{sec:annotation_predicate} and Section \ref{sec:related_elements_of_action_mention} present the annotation of predicate heads and relevant elements. Section \ref{sec:data_set} introduces the corpus employed in our work. Section \ref{sec:quality_control} discusses our strategies for controlling the quality of the annotated corpus.

\section{Related Work}
\label{sec:related-work}

The annotation of predicate heads is firmly rooted in the field of information extraction. To extract verbal or declarative information from sentences has received much attention over the past few decades \cite{Ref13}. In this section, before discussing related work about predicate head, the history of information is outlined as the background of predicate head annotation.

The first definition about information extract was presented by Schank et al. \cite{Ref14}, which proposed a linguistic structure known as conceptual dependency theory (CDT). It assumes that the main conceptualization of a clause is expressed by various concepts (e.g., actions and concrete nouns) and the relations between them. For every registered action, a conceptual structure is defined to express the relationship between relevant concepts. 

Frame theory is a popular framework used to represent the structure of semantic information \cite{Ref15}. A frame is a data structure of a predefined event. It contains a certain number of slots,  which can be filled by the semantic information about an event. For every frame, the number of slots and the type of fillers are manually defined according to a specific event. For example, a birthday party frame should contain hosts, guests and a birthday cake.

Lehnert et al. \cite{Ref16} presented a plot unit connectivity graph (PUCG) to represent the plot line of a story. In this graph, plot units are defined as conceptual elements referring to propositions or states, e.g., positive events, negative events, mental states, etc. A PUCG graph is composed of plot units linked by four types of relations: Motivation, Actualization, Termination and Equivalence.

For other early works, Rumelhart et al. \cite{Ref17} proposed a story grammar for extracting chronologies and plotting human narrative stories. Sagerc et al. \cite{Ref18} introduced a sublanguage to extract information from patient documents. 

For the past two decades, the message understanding conferences (MUC) and automatic content extraction (ACE)  are two important evaluation communities. In these communities, an event is defined as a template or frame with slots to be filled \cite{Ref4}. Slots are arguments of an event, e.g., action subjects, action objects, times, or locations. An event is triggered by a special word (anchor words, e.g., verbs). The task of recognizing events depends on finding all predefined arguments. 

Semantic role labeling (SRL) is a task of assigning semantic arguments to predefined predicates \cite{Ref19,Ref20}. Traditional semantic arguments include Agent, Patient, Instrument, etc. It also extracts adjunctive arguments such as Locative, Temporal, Manner, Cause, etc. In this task, predicates are not defined as the structural center of a sentence. Furthermore, not all predicates in a sentence are processed in the SRL task. The types of predicates are designated by the computational verb lexicon VerbNet \cite{Ref21}. In addition to verbal expressions, some nominal terms are adopted when they express verbal concepts \cite{Ref22}.

In the field of information extraction, a large-scale and high-quality evaluation corpus is a very valuable resource. At present, the well-known annotated corpora about predicates are FrameNet \cite{Ref23}, PropBank \cite{Ref24} and NomBank \cite{Ref25}. 

FrameNet labels the UK national corpus with frame theory \cite{Ref15}, in which a semantic framework is composed of a predicate (verb, partial noun, and adjective) and relevant semantic information. A frame is a schema about a specific event, which involves various participants. For example, a commerce frame may have several slots, e.g., buyer, goods, seller and money. In the FrameNet definition, all frames are predefined manually. The task to recognize a frame can be divided into two steps. First, detect whether a specific frame is mentioned in a text stream. Then, predefined semantic roles are identified according to the schema of the frame.

PropBank is based on the Penn TreeBank corpus, in which verbal propositions and their arguments are labeled. In the PropBank project, verbs are classified according to their verbal concepts. A verbal concept includes a set of allowable syntactic and semantic roles. These roles are numbered starting from zero. For example, the verb accept has a verbal sense of “take willingly”. This approach defines a role set with four arguments from Arg0 to Arg3 representing Acceptor, Thing accepted, Accepted-from and Attribute. The corpus is adopted to support the semantic role labeling task.

NomBank is an annotation project at New York University, which marks the noun predicate and semantic roles in Penn Treebank. The annotation specification of NomBank is following the PropBank project. Instead of annotating arguments for verbs, the goal of NomBank is to annotate arguments that co-occur with nouns in the PropBank Corpus. The corpus was developed by various assessment tasks, such as CoNLL2008 and CoNLL2009.

In the field Chinese NLP, three corpora are used to support verbal or declarative information extraction: Chinese Proposition Bank (CPB), Chinese NomBank \cite{Ref25} and Chinese FrameNet \cite{Ref26}. Definitions of these corpora are following the English versions. CPB is annotated based on the Chinese Penn Treebank corpus. Chinese NomBank extends the CPB annotation to include Chinese noun predicates. Chinese FrameNet is a Chinese dictionary based on framework semantic theory that follows the English FrameNet style.

In the field of Chinese information extraction, little work has been performed on Chinese predicate head recognition. Related works can be divided into two categories according to whether it defines the predicate head as the structural center of a sentence. 

In the first category, the predicate head is not seen as the structural center of a sentence. Researchers often adopt a definition the same as SRL, in which a sentence may contain multiple predicates \cite{Ref27,Ref28,Ref29}. In related works, predicates (or verbs) are given in advance, researchers mainly focus on recognizing relevant semantic roles for the given predicates.

In the second category, Chen et al. \cite{Ref30} constructed a probability-based recognition model. Sui et al. \cite{Ref31} proposed a predicate head–recognition method based on decision trees. Gong et al. proposed a method based on the combination of rules and statistics \cite{Ref32}. Li et al. \cite{Ref33} used the syntactic relationship between the subject and the predicate to identify predicates. At current, techniques to support these works are mainly based on rule-based or statistical methods. The main problem is that no generally accepted annotation and public evaluation data are available to support this line of work.

\section{Annotation of Predicate Head}
\label{sec:annotation_predicate}

In this section, issues about annotating predicate heads are discussed. The major challenge in annotating a predicate head is to guarantee its structural central role in a sentence. In Section \ref{sec:action_background}, several specifications are proposed to achieve the goal. Because determining the structure of predicate heads is the foundation to identify them, Section \ref{sec:patterns} outlines patterns of predicate heads.

\subsection{Specifications for Annotating Predicate Heads}
\label{sec:action_background}

Good annotation guidelines should satisfy properties such as consistency, neutrality and generality. In Chinese, these properties remain difficult to satisfy. Even an agreement on fundamental linguistic standards is difficult to reach among researchers. For example, four standard test corpora were used in the first international Chinese word segmentation Bakeoff \cite{Ref34}. A study has shown that, even for native Chinese speakers, consistency of segmentation is only 75\% \cite{Ref35}. In this annotation guideline, several strategies are proposed to decrease the complexity of annotation. They are helpful for annotators in making consistent decisions.

{\bf(1) Flattened structure}

The structure of a sentence is usually defined as a tree, in which a linguistic unit (e.g., a word, a phrase or a clause) is iteratively composed of smaller linguistic units. A tree structure is effective for representing the dependent relationships between linguistic units. The disadvantage of the tree structure is that, when the length of sentence is increased, the number of dependencies between words can exponentially increase. Iteratively generating a parsing tree is expensive in human labour and error prone.

Because a predicate head is the center of a sentence, it is defined in the top structure of a sentence. Therefore, a flattened structure better represents the semantic role of a predicate head in a sentence. Annotating the top-level structure has several advantages. First, the top-level structure contains a smaller number of linguistic units. The dependencies between them are clearer, which decreases the annotation complexity considerably. Second, focusing on the top-level structure eliminates the need to consider the height of a parsing tree or the hierarchical structure of a sentence. Third, the main meaning of a sentence is expressed at the top-level structure of a sentence, and annotating the top-level structure is in accord with the sentence semantic expression.

{\bf(2) Annotation of sentence boundaries}

The task of recognizing predicate heads is implemented at the sentence level. Before implementing the annotation task, documents should be segmented into sentences. Furthermore, a predicate head is the structural center of a sentence. Therefore, precisely identifying sentence boundaries is very important.

\begin{CJK}{UTF8}{gbsn}In Chinese, five punctuation marks are often used for sentence segmentation: “，” (comma), “。” (period), “；” (semicolon), “！” (exclamatory mark), and “？” (question mark). One common issue is that the use of commas (“，”) is ambiguous, which can be used to separate sentences and clauses \cite{Ref5}. In addition to the comma ambiguity problem, a Chinese sentence often contains several verbs with no morphological cue to distinguish them. As in the example shown in Section \ref{sec:challenges}, the sequence “陈某某持刀{\bf 捅刺}滕某某，用砖头多次{\bf 击打}其头部，并将其头部{\bf 撞向}地面，致其死亡。” contains four clauses, which have at least three verbal expressions: “捅刺”, “击打” and “撞向”. These expressions have similar syntactic roles. It is difficult to determine which one is the predicate head.\end{CJK}

\begin{CJK}{UTF8}{gbsn}
In this annotation corpus, sentence boundaries are manually annotated. To resolve the boundary ambiguity problem, the following rules are adopted to segment sentences.

1)	If a sentence is finished with an end mark (e.g., “。”, “；”, “！” or “？”) and contains no comma, it is processed as an independent sentence.

2)	A sentence may contain several clauses divided by commas. If a clause has a predicate head, it is segmented as a sentence.

3)	A complex sentence may have clauses linked by conjunctions. If each clause contains a predicate head, the complex sentence is divided from the conjunctions.

The above rules may segment a sentence into several smaller labeling units. This setting reduces the annotation complexity and ambiguity. It is helpful for annotators to make consistent and correct decisions.

{\bf(3) Conjunctions in a Sentence}

A conjunction can be used to link two linguistic units, e.g., sentences, clauses, phrases or words. If linked sentences or clauses have their own predicate heads, they (sentences or clauses) are segmented into individual sentences. However, people often use two verbal expressions with a conjunction in a sentence. For example, “两人发生争执并互相厮打” (They argue and hit each other). If a conjunction links two verbal expressions, it is unreasonable to label any one of them. 

To address the conjunction problem, the following rules are adopted to segment a sentence into smaller labeling units, which guarantees that one labeling unit contains only one predicate head.

1) If a conjunction is used in a complex sentence to connect two sentences or two clauses, the conjunction is used to segment the sentence.

2) If verbal expressions are directly linked by a conjunction, they are seen as a compound verbal expression annotated as predicate heads.

3) A conjunction may link two verb-noun phrases. This phenomenon is more complex. It will be discussed in Section \ref{sec:related_elements_of_action_mention} in details.

For example, “两人发生争执并互相厮打” contains two verbal expressions, “发生争执” and “互相厮打” linked by a conjunction, “并”. The compound verbal expression “发生争执并互相厮打” is annotated as a predicate head. This structure is widely used in Chinese, e.g., “驱车行驶” and “抓捕归案”. In fact, a conjunction usually makes no difference to the verbs’ syntactic or semantic meanings, e.g., “我们{\bf 驱车行驶}五十公里” and “我们{\bf 驱车和行驶}了五十公里”.

{\bf(4) Multi Verb-noun Phrases}

Several verb-noun phrases often occur in a Chinese sentence, Without morphological cues, it is difficult to distinguish them. The following two sentences are given to show this problem (each with two translations):

\vspace{0.25cm}
``王某拿出尖刀扎入张某左胸口''\\
\quad 1) Taking out a sharp knife, Wang Mou penetrated Zhang Mou's chest. \\
\quad 2) Wang Mou taken out a sharp and penetrated Zhang Mou's chest.
\vspace{0.25cm}

``黄某驾驶摩托车逃离现场''\\
\quad 1) Huang Mou drives a motorcycle for escaping from the scene. \\
\quad 2) Huang Mou escapes from the scene by driving a motorcycle.
\vspace{0.25cm}

As the above examples showing, without morphologic information in verb form, only depending on the context in a sentence, it is difficult to assess which one is better. This issue will be discussed in Section \ref{sec:related_elements_of_action_mention}.

{\bf(5) Modifiers and Aspects in a Sentence}

A word can have several types of modifier, e.g., adjective (“连跳” (continuous jump)), a noun (“{\bf 被告人}陈某某” (defendant Chen Moumou)), a quantifier (“{\bf 多次}击打” (repeatedly hit)) or a phrase (“{\bf 昨天收到的}茶叶”). Because the word and its modifier roles as a syntactical element, it is labeled as a whole mention. The word is marked as the head of the mention.

Labeling heads is helpful in identifying predicate heads or relevant elements. However, because no delimitation between Chinese words, it may leads to some ambiguity. For example, “连跳” means “continually jump”. This verb can be labeled as “连 (跳)”, where ``跳'' is a head. On the other hand, “弹跳” is a verb meaning “spring”. If it is segmented as “弹 (跳)” (“shot / jump”), the meaning of “spring” cannot be expressed. In this case, “跳” is not the head of “弹跳”.

In Chinese sentences, modifiers and aspect markers are usually difficult to distinguish from words. For example, terms “尖刀”, “宝刀”, “大刀” and “钢刀”  are registered as words in some lexicon. While, “弯刀” and “破刀” are referred as adjective-noun phrases. To simplify the problem, modifiers and aspect markers of a word are labeled as an independent linguistic element, in which the head (word) is included in a pair of parentheses, e.g., “多次(击打)” and “连(捅)”.

{\bf(6) Words in Predicate Heads}

New Chinese words are often generated by combining existing words or characters. Modern Chinese words can be roughly divided into 12 parts of speech: noun, verb, adjective, numeral, quantifier and pronoun, function word, adverb, preposition, conjunction, auxiliary, onomatopoetic word and interjection \cite{Ref36}. They play different roles for generating a verbal expression. Some words are rarely used to generate predicate heads, while some words must be considered carefully. According to the relationship between words and verbal expressions, we roughly group Chinese words into several categories. It is helpful for annotators in making decisions about predicate heads.
\begin{itemize}
\item[1)]The first category includes adjectives, onomatopoetic words and interjections. These words have an unconsolidated relation with verbs. Instead of used for generating predicate heads, they are mainly used to modify nouns or nominal phrases.

\item[2)]Nouns, pronouns and nominal phrases are often used as objects, subjects, times or locations. Distinguishing these words from predicate heads is also easy.

\item[3)]Numerals and quantifiers are easily distinguished from predicate heads too. These words can be used as modifiers of predicate heads to indicate the frequency of an action.

\item[4)]Adverbs, conjunctions and auxiliaries (or function words) are used to modify verbs with manner, place, time, frequency, degree, etc. These words usually have a coupling relation with predicate heads.

\item[5)]The fifth category is verbs. Verbs are usually used as predicate heads.
\end{itemize}

Among these five categories, adverbs, conjunctions and auxiliaries (or function words) are often difficult to distinguish from predicate heads. For example, in “我在公园” (I am in a park), the preposition “在” acts as a predicate head indicating, “I stay in a park”. In Chinese, depending on the context, a verb usually play different syntactic roles. For example, “他拿出尖刀” (take out a sharp knife) and “在拿出尖刀的时候” (when taking out a sharp knife), both “拿出” are verbs. However, the first is a predicate head. The second acts as a prepositional phrase.

In general, modifiers of predicate heads are adverbs, e.g., sentence “王某迅速取出一把尖刀” has an adverb “迅速” (quickly). Modifiers typically indicate the manner, place, time, frequency, degree, or level of certainty of an action. On the other hand, an aspect marker is commonly a function word (e.g., “得”, “了”, and “过” or negative word “不”, “否”, “非”, and “莫”) indicating the tense or aspect of an action. Compared to modifier words, aspect markers are rarely used independently to express semantic information. Aspect markers are usually used as auxiliaries of verbs.

When labeling Chinese predicate heads, modifiers are often difficult to distinguish from verb expressions. The problem for aspects is more serious. For example, in “拿下” (take down), “赶上” (catch up with) and “打碎” (break into pieces), “下”, “上” and “碎” express effects of the actions “拿” (take), “赶” (catch) and “打” (break). However, they usually register as words by many lexicons. This situation is even worse because more aspects can be simultaneously used in a phrase, for example, in “拿下来”, “赶上去” and “打碎了”. Because they share the same context, whether or not modifiers or aspects are used, we label verbal expressions with its modifiers, aspects and complements as predicate heads.

{\bf (7) Tagging symbols for predicate heads}

In the following, tagging symbols are introduced to support the annotation in this guideline.
\begin{itemize}
\item[1)] A labeling unit can be sentences or clauses. They are manually labeled.
\item[2)] A predicate head is enclosed with a pair of square brackets. In the left of each bracket pair, a subscript is used to indicate its type.
\item[3)] For every predicate head, an identification is given to indicate the pattern of the predicate head, which follows the element type.
\item[4)] The pattern of a predicate head is expressed by an “S”, “R”, “L”, “V” or “M” tag, which denotes the patterns of predicate heads. These patterns are discussed in Section \ref{sec:patterns}.
\end{itemize}

With the tagging symbols discussed above, the example in Section \ref{sec:challenges} is labeled as follows.

\vspace{0.25cm}

\noindent (1) \ 被告人陈某某因-家庭矛盾[$_{\text{PRE-S}}$迁怒]岳父滕某某。

\noindent (2) \ 2015年6月29日凌晨，陈某某[$_{\text{PRE-M}}$谎(称)]购买-房屋，

\noindent (3) \ 将-其[$_{\text{PRE-M}}$(骗)至]其新房南侧桥上，

\noindent (4) \ 两人[$_{\text{PRE-S}}$发生]争执

\noindent (5) \ 并[$_{\text{PRE-M}}$互相(厮打)]。

\noindent (6) \ 陈某某持-刀[$_{\text{PRE-S}}$捅刺]滕某某，

\noindent (7) \ 用-砖头[$_{\text{PRE-M}}$多次(击打)]其头部，

\noindent (8) \ 并将-其头部[$_{\text{PRE-M}}$(撞)向]地面，

\noindent (9) \  [$_{\text{PRE-S}}$致]其死亡。

\noindent (10) 陈某某驾驶电动三轮车[$_{\text{PRE-S}}$抛]尸至-大桥下的河中。
\vspace{0.25cm}

The clause “两人发生争执并互相厮打” is split into two labeling units (examples (4) and (5)). Take another example “我吃了一碗饭并且喝了一杯水”. The conjunction “并且” indicates that the verbs “吃” and “喝” have identical semantic roles. Which verb is a predicate head is difficult to determine. Therefore, we split the example into two sentences: “我吃了一碗饭” and “并且喝了一杯水”.

\subsection{Pattern of Predicate Heads}
\label{sec:patterns}

In Chinese language, compound words are widely used as verbal expressions. Because these words are usually generated by rules, they are impossible to list in a vocabulary (known as out-of-vocabulary words). In this section, according to the structure of predicate heads, predicate heads are classified into five patterns. In this corpus, patterns of predicate heads are manually annotated. The information is helpful for supporting the annotation.

In related works, Xue et al. \cite{Ref37} classified verb compounds into 7 categories, e.g., “verb compounds”, “verb (compound)+aspect marker”, “A-not-A (A-one-A)”, and “coordination with conjunctions”. To serve the the uniqueness property of predicate heads,  they are divided into five patterns. For example, “开发建设” and “开发和建设” are annotated as “coordinated verb compounds” and “coordination with conjunctions” in Xue et al. \cite{Ref37}. Because they are interchangeable in a sentence, both are annotated as a coordinated structure to guarantee the the uniqueness property. 

\vspace{0.25cm}
\noindent {\bf Patterns 1 : Singleton structure.} This structure denotes to predicate heads composed of a single transitive or intransitive verb without modifiers or aspect markers.
\vspace{0.25cm}

An “S” postfix is use to indicate the singleton structure. In this pattern, every predicate head is represented by a verb that is an entry in a given verbal lexicon. With this definition, four issues should be considered:

\begin{itemize}
\item[1)] In Chinese, many registered verbs contain characters indicating their tense or aspect. For example, “王某取得一把尖刀” and “王某取出一把尖刀”, where “得” and “出” can be seen as aspect markers indicating that a knife is already “obtained” or “ drawn”. However, “取得” is an entry in a verbal lexicon, but “取出” is not. Therefore, only “取得” belongs to this pattern, e.g., “王某[$_{\text{PRE-S}}$取得]一把尖刀”.

\item[2)]A compound verb may be formed by successive verbs. If this compound word has been registered in a lexicon, it is also considered as a singleton predicate head. For example, “反叛者们正在打砸抢” (Bands of rebels are beating, smashing and looting.). In Chinese, the word “打砸抢” can be segmented into three words “打/砸/抢” (beat/ smash/ loot). Because this word has been registered in a lexicon, it is annotated as a singleton predicate head.

\item[3)] Another case concerns Chinese intransitive verbs. Intransitive verbs are often composed of a verb and a noun. Examples include “下雨” (rain) and “下冰雹” (hail). Traditionally, the former is registered as a word in some lexicons, but the latter is not. Therefore, “下雨” is annotated as a predicate head. For the latter example, only “下” is annotated as a predicate head.

\item[4)] Verbalized nouns or verbalized adjectives are also annotated as singleton predicate heads. A verbalized adjective usually indicate a subject in some case. With this pattern, verbalized nouns are labeled as predicate heads, e.g., 我[$_{\text{PRE-S}}$开心], 我[$_{\text{PRE-S}}$喜欢], 我[$_{\text{PRE-S}}$漂亮], 我[$_{\text{PRE-S}}$幸福]. Many nouns and adjectives can be used as verbs,  e.g., ``左右[$_{\text{PRE-S}}$欲(刃)]相如'', ``[$_{\text{PRE-S}}$(红)透]半边天''. 
\end{itemize}

\vspace{0.25cm}
\noindent {\bf Pattern 2: Reduplicated structure. } A predicate head with a reduplicated structure contains reappearing verbs.
\vspace{0.25cm}

An “R” postfix is used to indicate a reduplicated structure,  e.g., “我们在操场[$_{\text{PRE-R}}$跑一跑]”. Chinese speakers often use reduplicated methods to generate compound words, e.g., AA, AAB, ABB, AABB, A里AB, A AB, and ABAB (e.g., “走走”, “跑一跑”, “洗洗澡”, “勾勾搭搭”, “慌里慌张”, and “比划比划”). Where, at least one verb is repeated. From the viewpoint of syntactic structure, any verb and its reduplicated versions are interchangeable without influence their roles as predicate heads.

Some word formation methods also use function words to generate reduplicated words, e.g., ``能不能'' and ``要不要''. Because Chinese words are multi-categorical in terms of part of speech, there annotation should depend on the context. For example, ``你[$_{\text{PRE-R}}$要不要]苹果'', and ``你[$_{\text{PRE-R}}$要不要(吃)]苹果'', the first ``要不要'' is a reduplicated predicate head. However, the latter ``要不要'' acts as a modifier.

\vspace{0.25cm}
\noindent  {\bf Pattern 3:Coordinated structure.} A predicate head can comprise coordinated verbs, which express relevant semantic meaning, e.g., “verb-resultative” and “verb-directional” compounds, or verbs with the same subcategorization frames.
\vspace{0.25cm}

An “L” postfix is used to annotate this pattern. In Chinese, synonymous or similar verbs are often collectively used to express an action, e.g., “驱车/ 行驶”, “开发/ 建设” and “抓捕/ 归案”. In man cases, a conjunction can be adopted to link coordinated verbs, e.g., “驱车/ 且/ 行驶”, “开发/ 和/ 建设”. The main difference between reduplicated and coordinated structures is that coordinated verbs consist of different verbs. On the other hand, reduplicated structure is composed of repeated verbs. These coordinated verbs are annotated as a whole predicate head:[$_{\text{PRE-L}}$驱车行驶], [$_{\text{PRE-L}}$开发建设], [$_{\text{PRE-L}}$抓捕归案].

Another type of predicate head is successive verbs representing a sequence of movements. Unlike verbs in coordinated structure which composes of synonymous or similar verbs, many successive verbs denote different actions. For example, “我去扭开水龙头” (I got to the stopcock and turn on it), where “去扭开” can be segmented as “去/ 扭开” (go/turn on). Resolving this problem depends on relevant elements. It is discussed in Section \ref{sec:related_elements_of_action_mention} in detail. In this example, it is labeled as ``[$_{\text{SUB-W}}$我][$_{\text{PRE-S}}$去][$_{\text{RAI-P}}$(扭开)水龙头]''. The motivation of this annotation is that ``去'' is the current action. However, ``扭开'' is the purpose of ``去''. For another example ``我去参加比赛'' (I go to the competition), it expresses the semantic meaning that ``I'm on the way to the competition'' (The competition hasn't started yet). Therefore, it is annotated as ``我[$_{\text{PRE-S}}$去]参加比赛''.

\vspace{0.25cm}
\noindent {\bf Pattern 4: Modified structure.} Verbs with modifiers, aspect markers and complements are labeled as modified structure predicate heads.
\vspace{0.25cm}

An “M” postfix is used to mark this pattern. In this case, the verb is annotated as the head of the predicate head. The verb is enclosed in parentheses. Therefore, the example “王某取出一把尖刀” is labeled as “王某[$_{\text{PRE-M}}$(取)出]一把尖刀”, where “出” is an aspect marker.

The pattern is helpful to simplify the annotating process because in Chinese many aspect markers or modifiers express semantic meaning about an action. Sometimes distinguishing aspect markers from a verb is difficult. Examples include ``我[$_{\text{PRE-M}}$(扭)开]电视机'' (I turn round the TV.), ``我[$_{\text{PRE-S}}$开]电视机'' (I open the TV), ``我[$_{\text{PRE-M}}$要(打开)]电视机'' (I will open the TV),  ``我[$_{\text{PRE-S}}$要]电视机'' (I want a TV).

\vspace{0.25cm}
\noindent {\bf Pattern 5:Specific structure.} These types include verbal expressions, e.g., proverbs, idioms, argots, allusions, etc.
\vspace{0.25cm}

An “V” postfix is used to mark this pattern, e.g., 王某某[$_{\text{PRE-V}}$心生不满] and 王某某[$_{\text{PRE-V}}$过河拆桥].

In summary, patterns of predicate heads are listed in Table \ref{tab:predicate_heads}.

\begin{table}[h]
\centering
		\caption{Patterns of Predicate Heads}
\label{tab:predicate_heads}
\begin{tabular}{|c|l|c|l|}
\hline
\textbf{No} & \multicolumn{1}{c|}{\textbf{Type}} & \textbf{Tag} & \multicolumn{1}{c|}{\textbf{Definition}}                                                                                                                \\ \hline
1           & singleton structure                & S            & \begin{tabular}[c]{@{}l@{}}A predicate head comprising a single transitive \\ or intransitive verb without modifiers or aspect \\ markers.\end{tabular} \\ \hline
2           & reduplicated structure              & R            & \begin{tabular}[c]{@{}l@{}}A predicate head containing at least a reappeared \\ verb.\end{tabular}                                                      \\ \hline
3           & coordinated structure              & L            & \begin{tabular}[c]{@{}l@{}}A predicate head comprising coordinated verbs,\\ which express relevant semantic meaning\end{tabular}                        \\ \hline
4           & modified structure                 & M            & \begin{tabular}[c]{@{}l@{}}A predicate head with modifiers, aspect markers \\ and complements.\end{tabular}                                             \\ \hline
5           & specific structure                 & V            & \begin{tabular}[c]{@{}l@{}}Other verbal expressions, e.g., proverbs,idioms,\\  argots, allusions, etc.\end{tabular}                                     \\ \hline
\end{tabular}
\end{table}

\section{Annotation of Relevant Elements}
\label{sec:related_elements_of_action_mention}

Predicate heads play a central role in representing and organizing syntactic and semantic information in sentences. To identify predicate heads, we must distinguish them from other linguistic roles in a sentence. Therefore, in addition to predicate heads {\bf (PRE)}, five linguistic elements are defined in this guideline: subject element{\bf  (SUB)}, temporal element{\bf  (TEM)}, locational element{\bf  (LOC)}, adverbial element{\bf  (ADV) }and complemental element {\bf (COM)}. In this paper, these elements are called predicate head–relevant elements (or relevant elements in short).

\subsection{Specifications for Annotating Relevant Elements}
\label{sec:background}

In the task of semantic role labeling \cite{Ref38}, agents are annotated as the main parameters of a verb or a predicate. However, in Chinese, the lack of morphology makes it very difficult to identify agents of predicates. Examples include “饭吃饱啦” and “水喝足啦”, where “饭” and “水” are receptor subjects. In these cases, without external knowledge, finding the agents of “吃” and “喝” is impossible. Therefore, instead of recognizing the agents of an action, subjects of predicate heads are annotated, which can be easily recognized.

Adverbial elements are words or phrases expressing the cause, manner or intent of predicate heads. Due to language ambiguity, distinguishing between the cause, manner or intent of an action is very difficult. For example, “He lit the fuses, and they ran for cover”, where “lit the fuses” can be a cause, an intention or a preparatory action of “ran for cover”. Therefore, combining them as adverbial elements can simplify the annotation task. Complemental elements are defined as phrases acted upon or caused by predicate heads. Based on this definition, the object of a predicate head is also a complemental element. The complemental element can also simplify the annotation problem. For example, direct and indirect objects need not be distinguished.

{\bf(1) Trigger of a Relevant Element}

A sentence often contains one or more prepositional phrases, e.g., “被告人陈某某因家庭矛盾迁怒岳父滕某某” (Due to some domestic issues, defendant Chen Moumou hates his father-in-law Teng Moumou). In this example, the preposition “因” is used to guide phrase “家庭矛盾”, which causes the action “迁怒”. Therefore, it is annotated as an adverbial element. The preposition is annotated as the trigger of a relevant element. In this guideline, triggers and relevant elements are enclosed in a square bracket and segmented by a “-” tag, e.g., “被告人陈某某 [因-家庭矛盾] 迁怒岳父滕某某”. A trigger can have a modifier or an aspect word (e.g., an adverb). In this case, the trigger is enclosed in parentheses, for example,[$_{\text{ADV-P}}$ 多次 (向)-被告人] 提出, [$_{\text{ADV-P}}$多次 (用)-砖头] 击打 and[$_{\text{TEM-W}}$  每 (当)-太阳落山的时候].

Many prepositions can be used as triggers to introduce a relevant element, e.g., “至”, “致”, “将”, “向”, “把”, “被”, “从”, and “对”. However, many verbs can be used as triggers too, e.g., ``用砖头多次击打''. In Chinese, distinguishing preposition-object and verb-object phrases is difficult, e.g., ``[向-被告人]提出'', ``[对-谢某甲等人]称'', ``[将-其(头部)]撞向'', ``[从-其家中]携带'', ``[把-雷蛟]送到'', ``[被-王某]杀害''. To produce consistent annotations, when annotating a relevant element, a “-P” postfix is used to represent that it is a preposition-object phrase or a verb-object phrase, e.g., [$_{\text{ADV-P}}$  用-砖头]多次击打 and[$_{\text{ADV-P}}$  向-头部]多次击打.

{\bf(2) Postfix of a Relevant Element}

In a sentence, a relevant element can be a word, a phrase or a clause. If a relevant element is a named entity or a word (which may have a modifier, e.g., adjective), it is labeled with a “-W” tag, e.g., “我打开” [$_{\text{COM-W}}$ 电视机]. If it has a modifier, the noun is enclosed in parentheses, e.g., [$_{\text{SUB-W }}$被告人(陈某某)]. For every relevant element, if it is not a “-P” or “-W” relevant element, it is marked with a “-C” postfix, e.g., 林某丙现妻迟某要求 [$_{\text{COM-C }}$ 栾少广管教栾某丙] and 我相信[$_{\text{COM-C }}$ 开门后会很失望].

{\bf(3)Successive Verbal Expressions in a Sentence}

Successive verbal expressions are widely used in Chinese. Making clear annotation rules is critical. Rules for annotating successive verbal expressions is listed as follows.

First, if a preparatory action is a verb-object phrase, it is labeled as an adverbial element. Here are two examples: “打开车门拿出一箱苹果” and “王某拿出尖刀扎入张某左胸口”. In the first case, “打开车门” and “拿出一箱苹果” are two successive verbs. “打开车门”” is the condition for implementing the action “拿出”. “一箱苹果” is acted upon by the action “拿出”. The sentence is labeled as [$_{\text{ADV-C}}$打开-车门] [$_{\text{PRE-M}}$(拿)出][$_{\text{COM-W}}$一箱 (苹果)]. In the second example, because “拿出尖刀” (taking out a sharp knife) is also a verb-object phrase, it is labeled as [$_{\text{SUB-W}}$王某][$_{\text{ADV-P}}$(拿)出-尖刀][$_{\text{PRE-M}}$(扎)入][$_{\text{COM-W}}$张某左(胸口)].

Third, if the first verb is not a verb-object phrase, it is labeled as a predicate head. For example, “我相信开门后会很失望”. In this example, the phrase “开门后会很失望” is the object of “相信” (believe). It is annotated as``[$_{\text{SUB-W}}$我][$_{\text{PRE-S}}$相信][$_{\text{COM-C}}$开门后会很失望]''. Here is another example: “二被告人商量决定寻找机会杀死张某甲”. This example contains four successive verbs “商量/ 决定/ 寻找机会/ 杀死”. We label this example as``[$_{\text{SUB-W}}$二被告人][$_{\text{PRE-S}}$商量][$_{\text{COM-C}}$决定寻找机会杀死张某甲]''.

Chinese words multi-categorical in terms of part of speech. Two successive verbal expressions may act as a verb-object phrase, for example, [我][引发][争论], [我][前往][商谈], [我][答应][去商谈] or [我][去][吃饭]. In these cases, the latter is labeled as a complemental element: [我][$_{\text{PRE-S}}$引发][$_{\text{COM-W}}$争论], [我][$_{\text{PRE-S}}$ 前往][$_{\text{COM-W}}$商谈], [我][$_{\text{PRE-S}}$ 答应][$_{\text{COM-P}}$去-商谈] or [我][$_{\text{PRE-S}}$ 去][$_{\text{COM-P}}$ 吃-饭].

{\bf(4) Auxiliaries in a Sentence}

Auxiliaries are used to express a possibility (e.g., “能够”, “可能”, and “可以”), a willingness (e.g., “愿意”, “想要”, “要想”, and “敢于”), a necessity (e.g., “应该”, “应当”, “得”, “该”, and “当”), or an assessment (e.g., “值得”, “便于”, “难于”, and “易于”). If auxiliary words are used as modifiers of a verb, they are labeled as parts of predicate heads, e.g.,[$_{\text{SUB-W}}$  我][$_{\text{PRE-M}}$ 可以(吃)掉][$_{\text{COM-W}}$ 这个 (苹果)]. However, if they express a verbal meaning, e.g., “要某人做某事” (ask somebody to do something), they are labeled as predicate heads, for example, [$_{\text{SUB-W}}$ 林某丙现妻 (迟某)][$_{\text{PRE-S}}$要求][$_{\text{COM-C}}$栾少广管教栾某丙].

{\bf(5) Idiomatic Usages in a Sentence}

Idiomatic usages (e.g., proverbs, idioms, argot or allusion, etc.) are an important part of the Chinese language. More than 18,000 idioms are registered in the Chinese idiom dictionary. These idioms are seen as a whole unit and annotated according to its semantic meaning. For example, [$_{\text{SUB-W}}$ 我][$_{\text{ADV-C}}$ 陪-你][$_{\text{PRE-S}}$到][$_{\text{LOC-W}}$天涯海角] and [$_{\text{SUB-W}}$我][$_{\text{ADV-P}}$陪-你][$_{\text{PRE-S}}$到][$_{\text{TEM-W}}$天荒地老].

{\bf(6) Quantifier in a Sentence}

The modifier of a verbal expression can be a quantifier, e.g., [$_{\text{ADV-P}}$ 用-砖头] [$_{\text{PRE-M}}$多次 (击打)][$_{\text{COM-W}}$其 (头部)], where “多次” means “many times”. A quantifier can appear anywhere in a sentence. Its type depends on the position in a sentence, e.g.,[$_{\text{ADV-P}}$ 多次 (用)-砖头][$_{\text{PRE-S}}$击打][$_{\text{COM-W}}$其 (头部)], [$_{\text{ADV-P}}$用-砖头][$_{\text{PRE-S}}$击打][$_{\text{COM-W}}$其 (头部)][$_{\text{COM-W}}$多次] and [$_{\text{ADV-P}}$用-砖头][$_{\text{PRE-S}}$击打][$_{\text{COM-W}}$多次]. These examples also indicate that the type of a relevant element is sensitive to its position in a sentence.

{\bf(7) Unclear sentences}

Many sentences can be annotated appropriately with this annotation guideline. Even so, a small number of sentences cannot be processed by the proposed annotation guideline. Unclear sentences are of two types. The first includes sentences written in the wrong way. Another type of unclear sentences is caused by idiomatic expressions. For example, the predicate head may be absent in some sentences. If a sentence is unclear, the whole sentence is annotated by a pair of square brackets with the type “UNC”.

{\bf(8) Tagging symbols for relevant elements}

Tagging symbols used for relevant elements are listed as follows.

\begin{itemize}
\item[1)]A relevant element is enclosed with a pair of square brackets.

\item[2)]For each relevant element, in the left of each bracket pair, a subscript is used to indicate its type.

\item[3)]A relevant element can be marked by a “-W”, “-P” or “-C” postfix, which indicate that the element is composed of a word, a phrase or a clause.

\item[4)]Triggers are segmented from relevant element by a “-” tag. Heads of relevant elements are included in parenthesis.

\end{itemize}

Based on the above tagging symbols, the example in Section \ref{sec:challenges} is completely labeled as follows.

\vspace{0.25cm}
\noindent (1) \ [$_{\text{SUB-W}}$被告人(陈某某)][$_{\text{ADV-P}}$因-家庭(矛盾)][$_{\text{PRE-S}}$迁怒][$_{\text{RAI-W}}$岳父(滕某某)]。

\noindent (2) \ [$_{\text{TEM-W}}$2015年6月29日凌晨]，[$_{\text{SUB-W}}$陈某某][$_{\text{PRE-M}}$谎(称)][$_{\text{COM-P}}$购买-房屋]，

\noindent (3) \ [$_{\text{ADV-P}}$将-其][$_{\text{PRE-M}}$(骗)至][$_{\text{LOC-W}}$其新房南侧(桥上)]，

\noindent (4) \ [$_{\text{SUB-W}}$两人][$_{\text{PRE-S}}$发生][$_{\text{COM-W}}$争执]

\noindent (5) \ 并[$_{\text{PRE-M}}$互相(厮打)]。

\noindent (6) \ [$_{\text{SUB-W}}$陈某某][$_{\text{ADV-P}}$持-刀][$_{\text{PRE-S}}$捅刺][$_{\text{COM-W}}$滕某某]，

\noindent (7) \ [$_{\text{ADV-P}}$用-砖头][$_{\text{PRE-M}}$多次(击打)][$_{\text{COM-W}}$其(头部)]，

\noindent (8) \ 并[$_{\text{ADV-P}}$将-其头部][$_{\text{PRE-M}}$(撞)向][$_{\text{COM-W}}$地面]，

\noindent (9) \  [$_{\text{PRE-S}}$致][$_{\text{COM-C}}$其死亡]。

\noindent (10) [$_{\text{SUB-W}}$陈某某][$_{\text{ADV-P}}$驾驶-电动三轮车][$_{\text{PRE-S}}$抛][$_{\text{COM-W}}$尸][$_{\text{COM-P}}$至-大桥下的河中]。
\vspace{0.25cm}

\subsection{Types of Relevant Elements}
\label{sec:relevant}

In this section, the proposed five elements are discussed: subject element, temporal element, locational element, adverbial element and complemental element. 

\vspace{0.25cm}
\noindent {\bf (1) Subject Element:} A subject element is a word or a phrase that control a predicate head.
\vspace{0.25cm}

A subject is defined as a word or a phrase which controls a verb in a sentence. It usually refer to an entity or a set of entities, for example, ``[$_{\text{SUB-W}}$王某]参加了这场竞赛'' and ``[$_{\text{SUB-W}}$王某等人]一起参加了这场竞赛''. Entities in a subject can be enumerated. The enumerated entities have the same semantic and syntactic information. In this case, they also labelled as a single mention, e.g., ``[$_{\text{SUB-W}}$王某, 赵某和张某]一起参加这场竞赛'' and [$_{\text{SUB-W}}$王某, 赵某和张某三人]一起参加这场竞赛''.  The subject can be absent in a sentence  subject, especially in imperative sentences, e.g., “请开门”.

When annotating a subject element, the following two issues should be considered. First, in a sentence, multiple entities in a subject can be positioned differently, possibly resulting in different linguistic roles, e.g., “王某下班后和赵某、张某二人一起参加了这场竞赛”. In this case, the first entity, “王某”, is the main actor. It is labeled as ``[$_{\text{SUB-W}}$王某][$_{\text{TEM-C}}$下班后][$_{\text{ADV-P}}$和-赵某、张某二人][$_{\text{PRE-M}}$一起(参加)了][$_{\text{COM-W}}$这场(竞赛)]'',  while the phrase “和赵某、张某二人” is a companion to conduct this action together.

The second issue is about the passive structure. Usually, in English, passive sentences can be identified using morphologies. Meanwhile, in Chinese, the lack of morphology makes passive sentences more challenging to recognize. For example, “饭吃饱啦”, “水喝足啦”, “饭” and “水” are receptor subjects. In these cases, identifying the agents of the actions “吃” and “喝” should depend on external knowledge. In this guideline, receptor subjects are also annotated as subject elements. This strategy simplifies the annotation task. 

\vspace{0.25cm}
\noindent {\bf (2) Temporal Element:} A temporal element indicates the time relevant to a predicate head.
\vspace{0.25cm}

In the field of information extraction, temporal expressions are semantic units conveying temporal information. They are usually handled as named entities and have received substantial attention\cite{Ref4,Ref39,Ref40,Ref25}. Compared with temporal expression recognition, the definition of a temporal element emphasizes its relationship with a predicate head. Only temporal expressions relevant to predicate heads are seen as the temporal element.

A temporal element can be expressed by a word, a phrase, a prepositional phrase (e.g., “在犯罪的时候”) or a clause (e.g., “我跑完步回来”). One rule for identifying a temporal mention is that it is referred to as a time point or a time period. Many words or phrases can express temporal semantic meanings such as “will” or “already” (e.g., “将要”, “马上”, and “曾经”). Instead of annotating them as temporal elements, these words or phrases are annotated as modifiers of a word. For example, “我晚上离开” (I’m leaving at evening.). “晚上” is a time period, which means “evening”. Because this word represents temporal information about the verb, the sentence is labeled as ``[$_{\text{SUB-W}}$我][$_{\text{TEM-W}}$晚上][$_{\text{PRE-S}}$离开]''.  In another example, “我马上离开” (I’m leaving right now), because “马上” is a modifier, it is labeled as  [$_{\text{SUB-W}}$我][$_{\text{PRE-M}}$马上(离开)]''.

A temporal expression can also be used as a subject, e.g., 2015 年 4 月 11 日是我的生日. In this case, the temporal expression is labeled as a subject element: [$_{\text{SUB-W}}$2015年4月11日][$_{\text{PRE-S}}$是][$_{\text{COM-W}}$我的(生日)].

\vspace{0.25cm}
\noindent {\bf (3)Locational Element:} A locational element is a locational expression relevant to a predicate head.
\vspace{0.25cm}

The issues about locational elements are the same as those for temporal elements. A locational element can also be expressed by a nominal phrase, a prepositional phrase or a clause. If a location is the subject of a sentence, it will be labeled as a subject mention. In this guideline, only locations relevant to a predicate head are labeled as locational elements.

\vspace{0.25cm}
\noindent {\bf (4) Adverbial Element:} An adverbial element indicates the cause, manner, intent or preparatory action relevant to a predicate head.
\vspace{0.25cm}

The reason for combining the cause, manner, intent or preparatory action of predicate heads into adverbial elements is that distinguishing them is difficult. For example, two successive actions may indicate a causal relationship, e.g., “He lit the fuse, and they ran for cover”. The phrases “用菜刀”, “用毛衣针” and “‘拿砖头” can be seen as the manner of implementing an action or as  preparatory actions of an action. Therefore, instead of trying to distinguish them, they are annotated as adverbial elements for consistency and simplicity.

Predicate heads may have one or several adverbial elements in a sentence. Some adverbial elements are verb-object phrases. Because Chinese lacks morphological cues, distinguishing them from predicate heads is difficult. The main rule for determining an adverbial element is the erasability principle, in which an adverb can be erased from a sentence without changing its main meaning. Another rule for identifying an adverbial mention is that the position of an adverbial element in a sentence is often before the predicate head. Examples include [$_{\text{SUB-W}}$杨守保][$_{\text{ADV-P}}$对-谢某甲等人][$_{\text{PRE-S}}$称][$_{\text{COM-C}}$杨某甲系因喝酒摔死的] or [$_{\text{SUB-W}}$杨某][$_{\text{ADV-P}}$像-一匹脱缰的野马][$_{\text{PRE-S}}$(冲)在][$_{\text{COM-W}}$前面].

\vspace{0.25cm}
\noindent {\bf (5) Complemental Element:} A complemental element is acted upon, results in or is influenced by a predicate head.
\vspace{0.25cm}

In the field of information extraction, a patient is often targeted as the argument of an action. For the same reason as was given for extracting agents, identifying a patient requires external knowledge beyond the information in a sentence. A complemental element refers to linguistic units that are acted upon, result in or are influenced by an occurrence of an action. These units often follow a predicate head.

Objects are the most important type of complemental elements. Two object types are commonly acknowledged in Chinese: direct and indirect. An indirect object is indirectly affected by an action. For example, in “我送你一个苹果” (I give you an apple.) “你” is the indirect object, and “苹果” is the direct object. In this guideline, direct and indirect objects are annotated as complemental elements. This scheme is helpful to ease the task of annotating elements, e.g.,  [$_{\text{SUB-W}}$我][$_{\text{PRE-S}}$送][$_{\text{COM-W}}$你][$_{\text{COM-W}}$一个(苹果)].. Another example is “我送一个苹果给你” (I give an apple to you.). In this guideline, “给” is a verb and “送一个苹果” is a verb-object phrase annotated as an adverbial element: [$_{\text{SUB-W}}$我][$_{\text{ADV-P}}$送-一个(苹果)][$_{\text{PRE-S}}$给][$_{\text{COM-W}}$你]. In another example, “我给你送一个苹果”, “给你” is a verb-object phrase acting as an adverbial element: [$_{\text{PRE-W}}$我][$_{\text{ADV-P}}$给-你][$_{\text{PRE-S}}$送][$_{\text{COM-W}}$一个(苹果)].

Here are two more examples: “我送给你一个苹果” and “我送你去机场”. In the first example, “送给” can be annotated as a predicate head with a reduplicated structure. Therefore, it is annotated as[$_{\text{SUB-W}}$我][$_{\text{PRE-R}}$送给][$_{\text{RAI-W}}$你][$_{\text{COM-W}}$一个(苹果)]''. In the second example, “送” expresses the meaning “accompany”. “你去机场” is the result of “送”. It is labeled as [$_{\text{SUB-W}}$我][$_{\text{PRE-S}}$送][$_{\text{COM-C}}$你去机场].

The definitions of these annotations are listed in Table \ref{tab:linguistic_role}.

\begin{table}[h]
\centering
		\caption{ Definitions of Annotated Linguistic Roles}
\label{tab:linguistic_role}
\begin{tabular}{|c|l|l|}
\hline
\textbf{Abbreviation} & \multicolumn{1}{c|}{\textbf{Type}} & \multicolumn{1}{c|}{\textbf{Definition}}                                                                                                             \\ \hline
\textbf{PRE}          & Predicate Head                     & \begin{tabular}[c]{@{}l@{}}A predicate head is a verbal expression \\ that acts as the structural and semantic \\ center of a sentence.\end{tabular} \\ \hline
\textbf{SUB}          & Subject Element                    & \begin{tabular}[c]{@{}l@{}}A subject element is a word or a phrase \\ that controls a predicate head.\end{tabular}                                   \\ \hline
\textbf{TEM}          & Temporal Element                   & \begin{tabular}[c]{@{}l@{}}A temporal element indicates the time \\ relevant to a predicate head.\end{tabular}                                       \\ \hline
\textbf{LOC}          & Locational Element                 & \begin{tabular}[c]{@{}l@{}}A locational element is a locational \\ expression relevant to a predicate head.\end{tabular}                             \\ \hline
\textbf{ADV}          & Adverbial Element                  & \begin{tabular}[c]{@{}l@{}}An adverbial element is the cause,\\ manner,intent or preparatory action \\ relevant to a predicate head.\end{tabular}    \\ \hline
\textbf{COM}          & Complemental Element               & \begin{tabular}[c]{@{}l@{}}A complemental element is acted upon, \\ results in or is influenced by a predicate\\ head.\end{tabular}                  \\ \hline
\end{tabular}
\end{table}

Temporal and locational elements are special cases of adverbial elements. They express time and location information about predicate heads. Because these elements are widely studied in the field of natural language processing, we annotate them separately from adverbial elements. One important rule for identifying relevant elements is that they must relate to predicate heads. For example, “他用昨天收到的茶叶泡了一杯茶”, is annotated as[$_{\text{SUB-W}}$他][$_{\text{ADV-W}}$用-昨天收到的 (茶叶)][$_{\text{PRE-M}}$(泡) 了][$_{\text{COM-W}}$一杯 (茶)]. In this example, “昨天” is related to “收到”. Because the predicate head of this sentence is “泡”, in this sentence, “昨天” is not annotated as a temporal element.

\subsection{Special Issues}
\label{sec:other_statements}

In this section, we discuss some special issues about predicate head–relevant elements.

\vspace{0.25cm}

{\bf (1) Ba-construction}: In this structure, Ba (“把”) is commonly used to introduce the object of an action. This word is called “subjective disposal”, and, in it, a speaker believes that the subject of a sentence has done something to an object. Because it expresses the intent of an action, this word is annotated as an adverbial element. In a Ba-construction, a verbal expression is often followed by a complemental element indicating the result of an action, e.g., [$_{\text{SUB-W}}$雨水][$_{\text{ADV-P}}$把-荷叶][$_{\text{PRE-M}}$(冲)得][$_{\text{COM-W}}$发亮], [$_{\text{SUB-W}}$王某][$_{\text{ADV-P}}$把-樊某][$_{\text{PRE-M}}$(移)至][$_{\text{LOC-W}}$本市南郊区云冈镇校尉屯村外(一土洞)].

\vspace{0.25cm}

{\bf (2) Bei-construction}: Bei is used in a passive sentence to introduce the agent of an action, e.g., “苹果被王某吃得干干净净”. In this structure, the receptor subject (“苹果”) is labeled as a subject element. The agent (“王某”) is labeled as an adverbial element, and the whole sentence is labeled as  [$_{\text{SUB-W}}$苹果][$_{\text{ADV-P}}$被-王某][$_{\text{PRE-M}}$(吃)得][$_{\text{COM-W}}$干干净净].

The Bei-construction may have other forms. 1) The agent can be omitted. Then, “被” is seen as the modifier of an action, e.g., [$_{\text{SUB-W}}$苹果][$_{\text{PRE-M}}$被(吃)得][$_{\text{COM-W}}$干干净净]. 2) A clause can follow the character “被”, e.g., [$_{\text{SUB-W}}$苹果][$_{\text{ADV-C}}$被-王某拿到屋内][$_{\text{PRE-M}}$(吃)得][$_{\text{COM-W}}$干干净净] (or [$_{\text{SUB-W}}$苹果][$_{\text{ADV-C}}$被-拿到屋内][$_{\text{PRE-M}}$(吃)得][$_{\text{COM-W}}$干干净净]. 3) Bei can occur together with a temporal element, e.g., [$_{\text{SUB-W}}$苹果][$_{\text{TEM-C}}$被-王某拿到屋内后][$_{\text{PRE-M}}$很快(吃)得][$_{\text{COM-W}}$干干净净].

\vspace{0.25cm}
{\bf (3) Position sensitive: }: The type of a relevant element is dependent on its position in a sentence. Because, in a sentence, syntactic information can be more accurately analyzed than semantic information, position information is helpful. Examples include[$_{\text{SUB-W}}$他][$_{\text{PRE-L}}$驱车行驶][$_{\text{COM-W}}$五十公里] and [$_{\text{SUB-W}}$他][$_{\text{ADV-P}}$驱车行驶-五十公里][$_{\text{PRE-S}}$到达][$_{\text{COM-W}}$收费站]. In the second sentence, a subject or a receptor subject of a sentence are annotated as subject elements without considering the object preposition problem.

\vspace{0.25cm}
{\bf (4)Priority between relevant mentions:}: A phrase may be annotated by several relevant element types at the same time. For example, a time or a location can be annotated as a subject element, an adverbial element or a complemental element, e.g., [$_{\text{SUB-W}}$2015年4月11日][$_{\text{PRE-S}}$是][$_{\text{COM-W}}$ 我的 (生日)]. The priority (from high to low) of annotating relevant elements is subject (temporal, locational) and attribute (complemental).

\section{Corpus}
\label{sec:data_set}

In our work, adjudication documents are chosen as the annotating corpus. These documents are written by judges based on statements of criminal facts. All adjudication documents have an officially predefined format. In China, the structure of adjudication documents is published by the supreme people’s court of China. Adjudication documents have five characteristics: normalization, innovativeness, publicity, legality and accuracy.

{\bf Legality}: Adjudication documents are highly professional documents written by judges. These legal documents are published by courts in accordance with legal functions and legal procedures.

{\bf Normalization}: To control the quality of adjudication documents and to normalize them, the supreme people’s court developed various specifications. All adjudication documents must follow technical and printing specifications, including font size, layout, numbers, etc. The normalization supports high performance information extraction.

{\bf Innovativeness}: The purpose of innovativeness is to avoid a dull format. When writing an adjudication document, the lack of innovativeness leads to a bad impression for readers, which diminishes the quality of a judgment. The innovativeness is interesting for a reader but becomes a challenge to the natural language process.

{\bf Publicness}: Adjudication documents embody rights and obligations. To ensure fairness, adjudication documents must be open to public access. This open access has the advantage that we can publicly obtain the data online.

{\bf Accuracy}: Adjudication documents must be written with neutral and objective sentences. Tendentiousness and emotionality are not allowed in adjudication documents. All sentences should be clear and accurate.

Adjudication documents are semi-structured data. Each adjudication document contains a paragraph describing the facts of a case. Each document begins with the specific phrase “人民检察院指控”. Then, the motive, process and result of a crime are written. The writing is required to be clear, accurate and objective. Therefore, adjudication documents are good resources for supporting the study of predicate head recognition.

\section{Quality Control}
\label{sec:quality_control}

Quality is the most important issue for an annotation project. The quality control runs throughout the whole process, which includes building annotation rules and annotating data.

\subsection{Building annotation rules}

Simplicity and unambiguity are two properties that are emphasized in our annotation guideline. An annotation guideline cannot be constructed in a single step. To support simplicity and unambiguity, we set annotation rules in steps. This section discusses the process we used to adjust the guideline.

In the first version of the annotation guideline, multiple predicate heads were allowed in a sentence. Types of relevant elements had more solid semantic information, e.g., causal factors, result factors and manner factors. The first version was used by four master’s students to annotate 20 documents. Each student annotated 5 documents. In the annotating process, students were required to record uncertain sentences. We found that, if multiple predicate heads were allowed in a sentence, the following problems resulted.

1) A multiple predicate head strategy may appear to avoid the burden of identifying the center of a sentence. However, determining which predicate head a relevant element belongs to is very difficult. In fact, this task increases the workload of annotating a corpus.

2) From the theory of cognitive psychology, a sentence with a center is easier to understand and remember (human beings find it difficult to simultaneously memorize several concepts). A center in a sentence also follows the dependency grammar theory, in which a root is defined in a syntactic tree. Therefore, multiple predicate heads in a sentence cause the structure of a sentence to be chaotic.

3) If multiple predicate heads are allowed in a sentence, the semantic role of predicate heads degenerates into verbs. Without the unique core role restriction, the definition recognizing predicate heads is similar to the POS-tagging task. No contribution is made to the NLP field.

4) Enabling multiple predicate heads in a sentence also limits their practical value. For example, to understand a story, the storyline should be represented as a chain of predicate heads. In a sentence, many verbal expressions are verbal clauses used as modifiers. These expressions are irrelevant to the main plot of the story.

Therefore, for the above reasons, in the revised annotation guideline, only one predicate head is allowed in a sentence. Several rules are used to support the single predicate head annotation. Annotators were required to revise the previous 20 documents with the new guideline. To examine the quality of the guideline, another 20 documents were annotated. Uncertain sentences were also collected for further processing.

For this revised annotation guideline, the main problem concerned the type of relevant elements. In the first version, in addition to the locational and temporal elements, factor types, such as causal, manner, object and resultant, were defined. However, in the annotating process, we found that these types were insufficient to cover predicate head–relevant elements. The solution is to increase the type of relevant elements. However, when the number of relevant element types increases, distinguishing them becomes very difficult. Then, instead of semantic types, we used types that are more “syntactic”. Therefore, in the third version, the adverbial and complemental element types were proposed.

Using the third annotation version, 11 master’s students and 3 undergraduate students were asked to annotate a new corpus. Each student was given 10 documents. One hundred forty documents were annotated. In the annotation process, uncertain sentences were also recorded.

The third version can better support the simplicity and unambiguity requirements. The annotation is easy to follow. The third version is the main framework of this guideline. Based on the third annotation version, in the rest of our work, several problems were revised, e.g., issues about idioms, successive verbal expressions, the Bei-construction, etc.

\subsection{Annotation process}

Using the final annotation guideline, 21 master’s students were recruited to conduct the annotating work. All students were first required to learn the guideline. Then, these students received a training program. All the ambiguous sentences are discussed among the students. When all the data were annotated, the students and the annotated corpus were divided into two groups. The annotated corpus was mutually cross-checked between the groups. If necessary, a meeting was conducted to resolve the questions raised by annotators. We iterated this process until an agreement was reached.


\section{Acknowledgements}
\label{sec:size}

This work was supported in part by the National Natural Science Foundation of China through the Joint Funds under Grant U1836205, in part by the National Natural Science Foundation of China through the Major Research Program under Grant 91746116, in part by the National Natural Science Foundation of China under Grant 62066007 and Grant 62066008, in part by the Major Special Science and Technology Projects of Guizhou Province under Grant [2017]3002, and in part by the Key Projects of Science and Technology of Guizhou Province under Grant [2020] 1Z055.
\end{CJK}


%
%



\end{document}